\documentclass[letterpaper]{article} % DO NOT CHANGE THIS
\usepackage{aaai25}  % DO NOT CHANGE THIS
\usepackage{times}  % DO NOT CHANGE THIS
\usepackage{helvet}  % DO NOT CHANGE THIS
\usepackage{courier}  % DO NOT CHANGE THIS
\usepackage[hyphens]{url}  % DO NOT CHANGE THIS
\usepackage{graphicx} % DO NOT CHANGE THIS
\usepackage{natbib}  % DO NOT CHANGE THIS AND DO NOT ADD ANY OPTIONS TO IT
\usepackage{caption} % DO NOT CHANGE THIS AND DO NOT ADD ANY OPTIONS TO IT
\usepackage{graphicx}
\usepackage{multirow}
\usepackage{amssymb}
\usepackage{amsfonts}
\usepackage{algorithm}
\usepackage{algorithmic}
\usepackage{{amsmath}}
\usepackage{booktabs}

\usepackage{newfloat}
\usepackage{colortbl}
\usepackage{listings}
\usepackage{amssymb}
\DeclareCaptionStyle{ruled}{labelfont=normalfont,labelsep=colon,strut=off} % DO NOT CHANGE THIS
\floatstyle{ruled}
\newfloat{listing}{tb}{lst}{}
\floatname{listing}{Listing}
\title{Mixture-of-Attack-Experts with Class Regularization for Unified Physical-Digital Face Attack Detection}
\author{
    Shunxin Chen\textsuperscript{\rm 1,2,3}\equalcontrib, Ajian Liu\textsuperscript{\rm 4,5}\equalcontrib, Junze Zheng\textsuperscript{\rm 5}, Jun Wan\textsuperscript{\rm 4,5,8}\thanks{Corresponding author}, Kailai Peng\textsuperscript{\rm 6}, \\
    Sergio Escalera\textsuperscript{\rm 7}, Zhen Lei\textsuperscript{\rm 4,5,8,9}    
}
\affiliations{
    \textsuperscript{\rm 1}Nanjing University of Posts and Telecommunications, Nanjing, China\\
    \textsuperscript{\rm 2}Nanjing Artificial Intelligence Research of IA, Nanjing, China\\
    \textsuperscript{\rm 3}University of Chinese Academy of Sciences, Nanjing, China\\
    \textsuperscript{\rm 4}MAIS, Institute of Automation, Chinese Academy of Sciences, Beijing, China\\
    \textsuperscript{\rm 5}Macau University of Science and Technology (MUST), Macau, China\\
    \textsuperscript{\rm 6}Purple Mountain Laboratory, China\\
    \textsuperscript{\rm 7}Computer Vision Center (CVC), Barcelona, Catalonia, Spain\\
    \textsuperscript{\rm 8}School of Artificial Intelligence, University of Chinese Academy of Sciences, China\\
    \textsuperscript{\rm 9}CAIR, HKISI, Chinese Academy of Sciences, Hong Kong, China\\
    chenshunxin23@mails.ucas.ac.cn, \{ajian.liu,jun.wan\}@ia.ac.cn
}

\usepackage{bibentry}
\begin{document}

\maketitle

\begin{abstract}
Unified detection of digital and physical attacks in facial recognition systems has become a focal point of research in recent years. However, current multi-modal methods typically ignore the intra-class and inter-class variability across different types of attacks, leading to degraded performance.% To address this limitation, we propose the Fine-Grained MoE with Class-Aware Regularization CLIP framework (FG-MoE-CLIP-CAR), which
To address this limitation, we propose MoAE-CR, a framework that effectively leverages class-aware information for improved attack detection. Our improvements manifest at two levels, i.e., the \textit{feature} and \textit{loss} level.
% To address this limitation, we propose the Mixture-of-Attack-Experts with Class Regularization (MoAE-CR) framework, which incorporates key improvements at both the \textit{feature} and \textit{loss} levels.
% MoAE-CR, a framework that effectively leverages these informative clues by incorporating key improvements at both the \textit{feature} and \textit{loss} levels.
% that effectively leverages these class-aware clues, namely the 
% incorporates key improvements at both the \textit{feature} and \textit{loss} levels. 
\textbf{At the feature level}, we propose Mixture-of-Attack-Experts (MoAEs) to capture more subtle differences among various types of fake faces. 
\textbf{At the loss level}, we introduce Class Regularization (CR) through the Disentanglement Module (DM) and the Cluster Distillation Module (CDM). 
The DM enhances class separability by increasing the distance between the centers of live and fake face classes. 
However, center-to-center constraints alone are insufficient to ensure distinctive representations for individual features. Thus, we propose the CDM to further cluster features around their class centers while maintaining separation from other classes. 
Moreover, specific attacks that significantly deviate from common attack patterns are often overlooked. 
To address this issue, our distance calculation prioritizes more distant features. 
Extensive experiments on two unified physical-digital attack datasets demonstrate the State-of-The-Art (SoTA) performance of the proposed method.
\end{abstract}

% Uncomment the following to link to your code, datasets, an extended version, or similar.

\section{Introduction}
\begin{figure}[t]
    \centering
    \includegraphics[width=\linewidth]{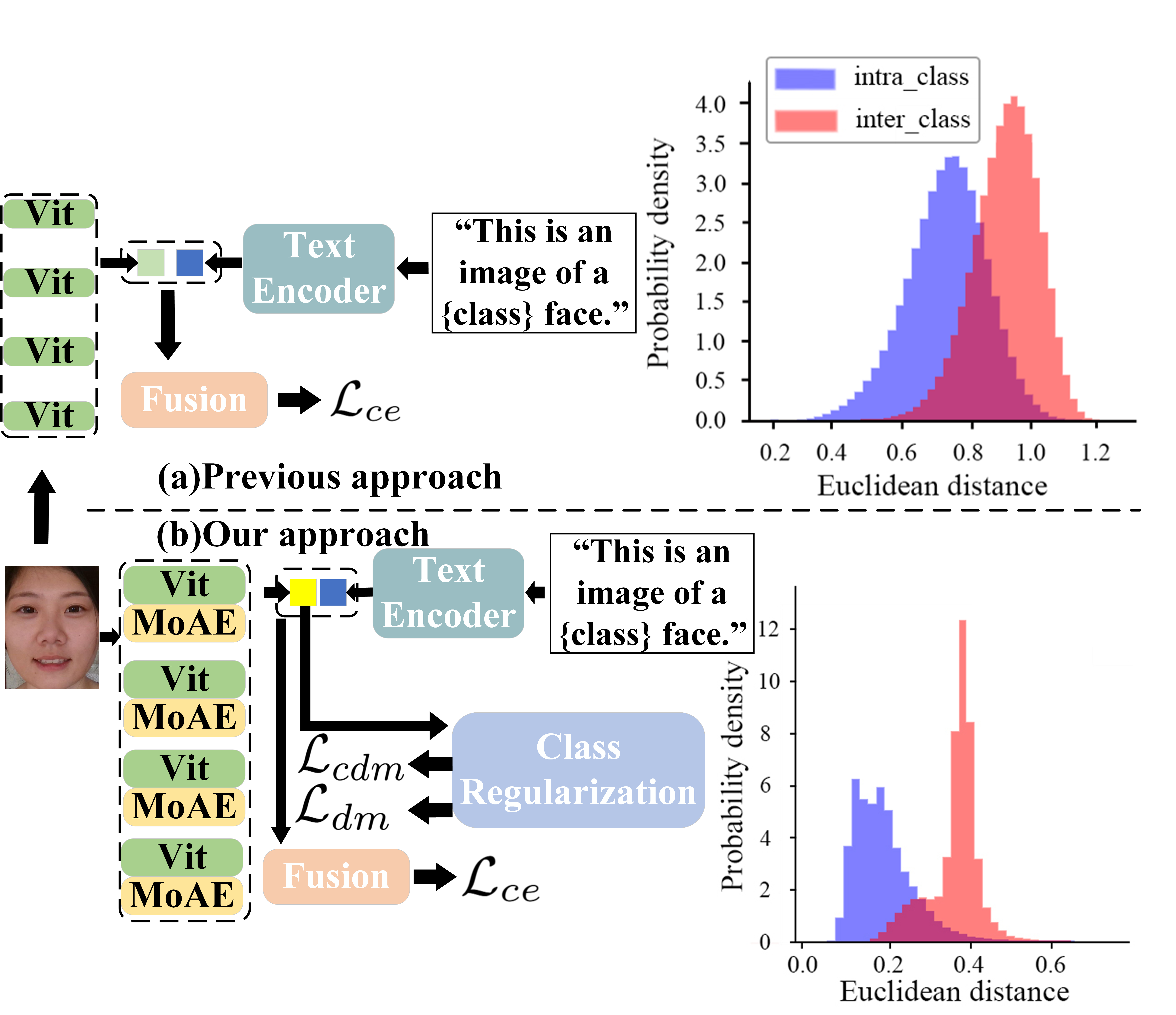} % 替换为图片的路径
    \caption{Comparison with existing methods.  Greater overlap in histograms indicates poor class separation.  (a) Previous methods focus on feature mining but overlook intra-class and inter-class variations.  (b) Our method refines features and enforces constraints, achieving a more distinct and separable feature space.
    }
    \label{fig111}
\end{figure}
% Facial recognition systems are increasingly prevalent in real-world applications, encompassing domains such as security, surveillance, and personal device authentication. Despite their widespread adoption, these systems 
Facial recognition systems remain susceptible to a variety of attacks, broadly classified into physical and digital attacks. Each category comprises distinct types of attacks: physical attacks include print attacks~\cite{zhang2020casia}, replay attacks, and mask attacks~\cite{liu2022contrastive,fang2023surveillance}, whereas digital attacks encompass methods~\cite{roessler2019faceforensicspp} such as StyleGAN, FaceSwap, Deepfakes, and NeuralTextures. Research on physical attack detection~\cite{liu2018learning,zhang2019dataset,yu2020fas,cai2020drl,liu2021face} often involves the design of specialized networks to automatically extract spoofing cues and deceptive features from multiple modalities. In the realm of deepfake detection, numerous studies~\cite{fei2022learning,qian2020thinking} leverage the spatial rich model, frequency details, and the relationships between facial action units to distinguish between genuine and fake faces. However, these methods cannot effectively address different types of attacks in different categories. % Despite the notable detection rates achieved by various methods and their respective advancements, real-world attacks frequently exhibit a combination of physical and digital elements, thereby rendering single detection methodologies ineffective. Given that the exact nature of a facial attack may not be known a priori, it is imperative to develop a general model capable of safeguarding facial recognition systems against any type of attack.
\begin{figure}[t]
    \centering
    \includegraphics[width=\linewidth]{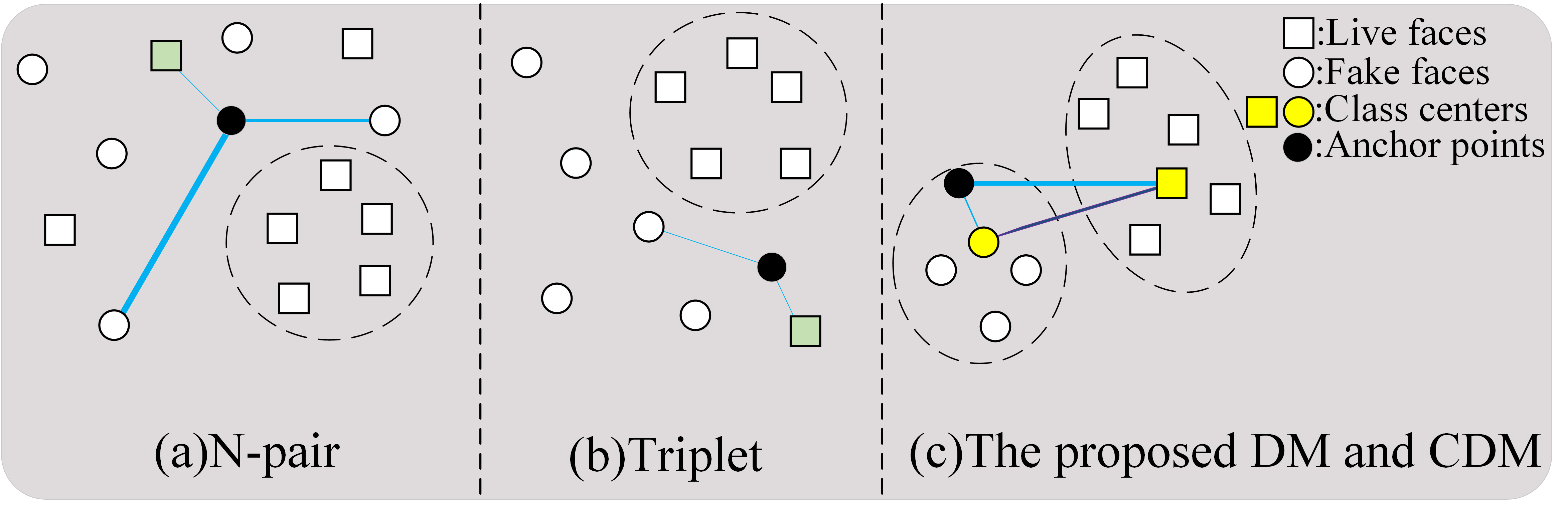} % 替换为图片的路径
    \caption{
    Comparison of popular class constraint methods and our approach. Small nodes represent the features of batch data. The connections defined by the loss are represented by edges, with thicker edges indicating larger gradients. (a) The N-pair loss reflects the hardness of the data but does not utilize all the data in the batch. (b) The triplet loss does not account for data hardness. The aggressive pushing mechanisms utilized by both (a) and (b) can lead to unintended class separation. Such forceful displacement may cause certain points, particularly green points, to diverge from their respective class clusters. (c) Our method considers all data in the batch, processes them with class centers, and simultaneously avoids class separation phenomena.
    % Comparison of class constraint methods. The connections defined by the loss are represented by edges, with thicker edges indicating larger gradients. (a) The N-pair loss reflects the hardness of the data but does not utilize all the data in the batch. (b) The triplet loss does not account for data hardness. The aggressive pushing mechanisms utilized by both (a) and (b) can lead to unintended class separation. Such forceful displacement may cause certain points, particularly green points, to diverge from their respective class clusters. (c) Our method considers all data in the batch, processes them with class centers, and simultaneously avoids class separation phenomena.
    }
    \label{fig222}
\end{figure}

Previous approaches~\cite{fang2024unified} seek to achieve classification by exploring a comprehensive feature space but do not account for the large intra-class differences and small inter-class differences present in physical and digital attack data. The inherent characteristics of the data make it extremely challenging to identify such a space, and neglecting these characteristics leads to suboptimal performance, as shown in Fig.~\ref{fig111}. Common techniques for implementing class regularization include triplet loss~\cite{schroff2015facenet,wang2014learning} and N-pair loss~\cite{sohn2016improved}. As illustrated in Fig.~\ref{fig222}, N-pair loss connects an anchor point to a single positive data point and multiple negative data points, pulling the positive point closer to the anchor while pushing the negative points away, with consideration of their hardness. However, N-pair loss does not fully utilize the entire data batch, as it samples an equal number of points from each negative class, which may result in the exclusion of informative samples during training. On the other hand, triplet loss fails to adequately account for the difficulty of the data, resulting in limited sensitivity when processing distant features. Additionally, the direct pull-and-push mechanisms in both triplet loss and N-pair loss pose challenges when dealing with classes that have enveloping relationships, making it difficult for dispersed classes to cluster effectively, thereby limiting the model’s ability to learn a unified feature representation.

In this paper, we propose the Mixture-of-Attack-Experts with Class Regularization (MoAE-CR) framework, which incorporates SoftMoE~\cite{puigcerver2023sparse} into the image encoder of CLIP~\cite{radford2021learning} to enable more refined processing at the feature level. We further refined and proposed MoAE, enabling it to process features from multiple perspectives with greater granularity. Additionally, we introduce two novel constraint modules: the Disentanglement Module (DM) and the Cluster Distill Module (CDM). These modules account for all data within a batch during computation and employ a relational matrix to prevent class separation caused by simple pushing mechanisms. DM enhances the separation between these classes to address the challenge of small inter-class differences, particularly in distinguishing between real and fake faces. Meanwhile, CDM promotes the clustering of features around their respective class centers while maintaining separation from other class centers. Furthermore, distance is utilized as a constraint reference to mitigate the model's overlooking rare attacks. In summary, the main contributions of this paper are as follows:
\begin{itemize}
    \item We propose a novel MoAE-CR framework, which incorporates MoAE and two regularization modules, DM and CDM. It demonstrates undeniable advantages over SoTA methods on two unified physical-digital attack datasets.
    \item At the feature level, we integrate SoftMoE into the image encoder of CLIP. To enable finer feature processing and capture the nuances of various attack types, MoAE enhances SoftMoE through the application of multi-head attention mechanisms.
    \item At the loss level, we utilize two constraint modules, DM and CDM. These modules ensure that live and fake faces exhibit greater intra-class aggregation and inter-class separation. In processing live and fake faces, all data within a batch is considered, with careful attention to the impact of distances. By assigning larger gradients to more distant features, we more effectively address attacks with skewed feature distributions.
\end{itemize}

\section{Related Works}
\subsection{Face Anti-Spoofing}
Face anti-spoofing is a technique designed to identify whether a face captured by sensors is genuine or a presentation attack, i.e., prints~\cite{zhang2020casia}, video replays, or 3D mask attacks~\cite{liu2022contrastive,fang2023surveillance}. % Early FAS was mainly used to distinguish by capturing the texture difference between true and false faces, such as LBP~\cite{maatta2011face} and HOG~\cite{komulainen2013context}. 
With the advancement of deep learning, researchers~\cite{liu2018learning,yu2020fas,cai2020drl} have developed specialized networks that automatically extract spoofing cues. However, these algorithms suffer from performance degradation when facing unknown domains. To address this issue, recent methods have employed DA-based techniques~\cite{liu2022source,yue2023cyclically,liu2024source} and DG-based approaches \cite{DBLP:journals/tifs/ZhengLWWMLDW24,10448479,liu2023towards,cai2024towards,liu2024cfplfas,liu2024moeit} aim to learn domain-invariant features across multiple source domains. Also, incremental learning (IL) methods~\cite{siw-mv2,wang2024multi} are considered to tackle the catastrophic forgetting problem in the context of domain discontinuity in FAS. With the increasing advancement of physical presentation mediums, an increasing number of algorithms are mining complementary information from visible light, depth map, and near-infrared modes to identify spoofing clues, including multi-modal fusion~\cite{zhang2020casia,george2019biometric}, cross modal transformation~\cite{george2021cross,liu2021face}, flexible modal~\cite{liu2023ma,liu2023fm,yu2023flexible,yu2023visual,10744492,10.1145/3664647.3680856}, and missing modality~\cite{lin2024suppress,zheng2024learning,DBLP:conf/icb/LiYLSGSKK24}.

\subsection{Face Forgery Detection}
Digital attack detection~\cite{zhao2021multi,LMM-forgery-detection} aims to distinguish authentic images from digitally manipulated facial artifacts, or diffusion generated video. Numerous endeavors have been undertaken to enhance the efficacy of Digital attack detection techniques~\cite{nguyen2019capsule,tolosana2020deepfakes}. In initial studies~\cite{rossler2019faceforensics++}, image classification backbones were employed to extract features from isolated facial images, facilitating binary classification. 
With the increasing visual realism of forged faces, recent efforts focus on identifying more reliable forgery patterns, including noise statistics, local textures, and frequency information. Zhao et al.~\cite{zhao2021multi} introduced a texture enhancement block in shallow layers to extract and enhance texture features by applying average pooling to filter out texture details from feature maps and subtracting the result from the original image. For both cnn-synthesized and image editing forgery domains, HiFi-IFDL~\cite{guo2023hierarchical} and HiFi-Net++~\cite{CLIP-forgery-detection} formulate the image forgery detection and localization (IFDL) as a hierarchical fine-grained classification problem, and classify the individual forgery method of given images via predicting the entire hierarchical path. DD-VQA~\cite{zhang2025common} extends deepfake detection from a conventional binary classification to a VQA task. 
% Other researchers~\cite{frank2020leveraging} revealed that the core reason for the different spectrums is the upsampling operation, which also hinted at the structural problem of GAN mapping from low-dimensional latent space to high-dimensional space. Several other works~\cite{zi2020wilddeepfake,dong2022protecting} have integrated attention mechanisms to bolster the differentiation between authentic and manipulated images.

\subsection{Physical-Digital Attack Detection}
Recent studies have made significant strides in the detection of face fraud and forgery. A pioneering benchmark for detecting face fraud and forgery was established, integrating visual and physiological rPPG signals to address issues of generalization~\cite{yu2024benchmarking}. Additionally, a comprehensive analysis of 25 documented attack types introduced a method that utilizes a multi-task learning framework alongside k-means enhancement techniques to differentiate between genuine identities and various attacks~\cite{deb2023unified}. La-SoftMoE~\cite{zou2024softmoe} leverages re-weighted SoftMoE with linear attention, achieving satisfactory performance on tasks with sparse feature distributions. But its generalizaiton would not meet practical requirements. Further contributions include the development of a new benchmark using existing physical and digital attack datasets, employing reconstruction learning for detection~\cite{cao2024towards}. However, these studies have not investigated unified attack detection based on ID consistency. % This article primarily utilizes the UniAttackData~\cite{fang2024unified}.
\begin{figure*}[t]
    \centering
    \includegraphics[width=\textwidth]{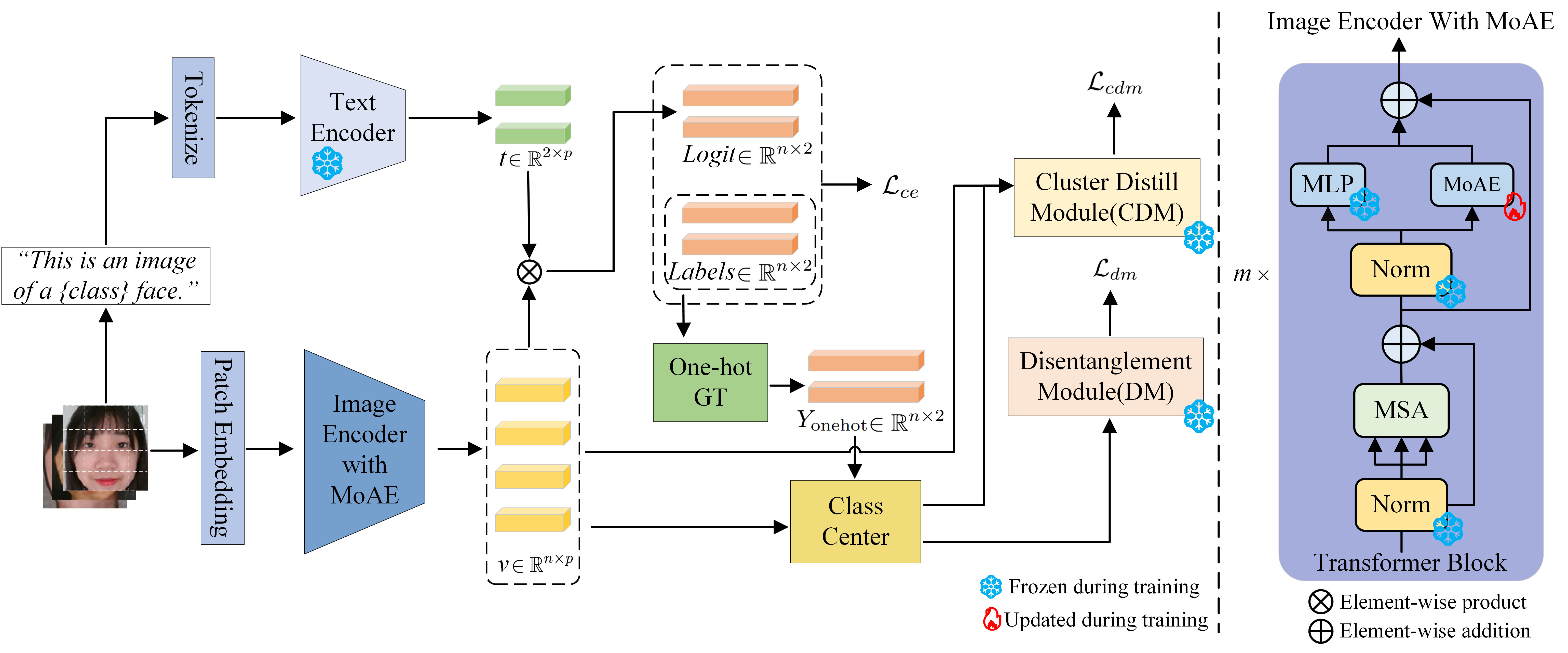} % 替换为图片的路径
    \caption{Our proposed MoAE-CR framework. This article primarily utilizes the Uniattack Datasetoder, and is designed to adapt to joint physical and digital attack tasks through contributions at two levels: (1) The image encoder incorporates MoAEs, which are composed of $m$ Transformer Blocks. Our MoAEs facilitate more nuanced learning from multiple perspectives, resulting in superior feature representation. (2) Two constraint modules: the Disentanglement Module (DM) and the Cluster Distill Module (CDM). These modules maximize intra-class cohesion and inter-class separation between live and fake faces.}
    \label{fig333}
\end{figure*}

\section{Method}
Our approach is based on a vision-language model, as illustrated in Fig.~\ref{fig333}. It involves the adjustment of features through an enhanced MoAE, while introducing a CR module at the loss layer to facilitate intra-class aggregation and inter-class separation. Subsequently, we will provide a detailed introduction to the proposed MoAE-CR.

\subsection{Preliminary}
In this paper, we utilize large-scale Vision-Language Models~\cite{jiang2024effectiveness,huang2024empiricalstudyllama3quantization} (VLMs) like CLIP (Contrastive Language-Image Pre-Training)~\cite{radford2021learning} based on contrastive learning to dynamically adjust classifier weights using textual features. CLIP integrates text and image encoders: a Transformer-based model converts text into fixed-size vectors, while convolutional neural networks (ResNet or Vision Transformer) convert images into vectors of the same dimensionality. Through contrastive learning, the model aligns image and text representations within a shared embedding space. The training objective is to minimize the cosine distance for positive pairs and maximize it for negative pairs using a symmetric cross-entropy loss function.

In our model, we keep the cross entropy of the similarity between images and texts in the CLIP model. The formula is shown as Eq. 1, where $S$ is the similarity matrix, and the element $S_{i,j}$ of the similarity matrix $S$ denotes the similarity between the $i^{th}$ image embedding and the $j^{th}$ text embedding.
\begin{small}
\begin{equation}\!\mathcal{L}_{ce}\!=\!-\!\frac{1}{2N}\!\sum_{i=1}^{N}\!\left(\!\log\!\frac{\exp(S_{i,i})}{\sum_{j\!=\!1}^{N}\!\exp(S_{i,j})}\!+\!\log\!\frac{\exp(S_{i,i})}{\sum_{j\!=\!1}^{N}\!\exp(S_{j,i})}\!\right)\!.\end{equation}
\end{small}

In our task, the text in CLIP is limited to two categories: live and fake, preventing the exploration of specific attack types. To address data sparsity, we employ Soft Mixture of Experts (Soft MoEs). Unlike sparse and discrete routing mechanisms that assign tokens to experts definitively, Soft MoEs use a flexible approach by mixing tokens to handle features. This involves calculating multiple weighted averages of all tokens, with weights determined by both tokens and experts. These weighted averages are then processed by their respective experts. Soft MoEs show exceptional performance in handling sparse tasks in visual recognition.

\subsection{Mixture-of-Attack-Experts}
% \subsection{Fine-Grained MoE CLIP with Class-Aware Regularization}
% \subsubsection{Fine-Grained MoE.}
As illustrated in Fig.~\ref{fig333}, the MoAE module is integrated into the image encoder, where it operates in parallel with the MLP within the transformer block, and the results are subsequently combined. Given the subtle differences among various types of deception attacks in our task, we propose the MoAE module. In our enhanced MoAE, the primary improvements involve the introduction of multi-head attention~\cite{katharopoulos2020transformers} to augment the model's representational capacity and learning ability, as well as the parallel processing of representations from different heads within the expert networks.
Multiple experts can capture various attack features, and the use of soft routing allows for the weighted aggregation of features processed by all experts. The addition of multiple heads enables different branches to learn distinct attack traces, thereby enhancing the effectiveness of the anti-spoofing task.

Specifically, given an input $x\in\mathbb{R}^{n\times p\times d}$ where $n$ denotes the batch size set to 32, $p$ represents the sequence length, and $d$ is the feature dimension, we first perform a linear transformation to obtain the query vector $q$, the key vector $k$, and the value vector $v$. Each MoE layer uses a set of $m$ expert functions applied on individual tokens, namely 
$\left\{f_{i}: \mathbb{R}^{d} \rightarrow \mathbb{R}^{d}\right\}_{1: n}$. These vectors are divided into multiple heads. Subsequently, the attention scores for each head are computed and normalized, followed by applying the softmax function to the attention scores to obtain the attention weights, the attention output $P$ is obtained by applying weighted summation of the value vectors $v$ using the attention weights:
\begin{equation}P=\mathrm{softmax}\left(\frac{q\cdot k^T}{\sqrt{d_h}}\right)\cdot v,\end{equation}
where $d_h=\frac dh$, $h$ is the number of heads.

In the improvement of the expert network, we process the output of each attention head separately through the expert network, allowing each head to focus on different feature representations. Specifically, given the output of each head, we need to reshape it to match the input format of the expert network. Subsequently, these reshaped outputs are processed through the expert network:

\begin{equation}
{\mathrm{Y}}_{i}=f_{\lfloor i / p\rfloor}\left({P}_{i}\right) .
\end{equation}

Finally, the output $Y$ of the expert network is reshaped back to its original form.

\subsection{Class Regularization}
The Class Regularization module comprises two constraint components: the Disentanglement Module (DM) and the Cluster Distillation Module (CDM). The DM increases the distance between the centers of real and fake face classes, thereby facilitating the separation of these classes and mitigating any potential dependencies between them. The CDM further refines feature clustering around their respective class centers while maximizing separation from centers of other classes, thereby optimizing class-specific characteristics. Additionally, to address the unique features found in combined physical and digital attacks, we prioritize features that are more distant during distance calculations.
\subsubsection{Disentanglement Module.}
\begin{figure*}[t]
    \centering
    \includegraphics[width=\textwidth]{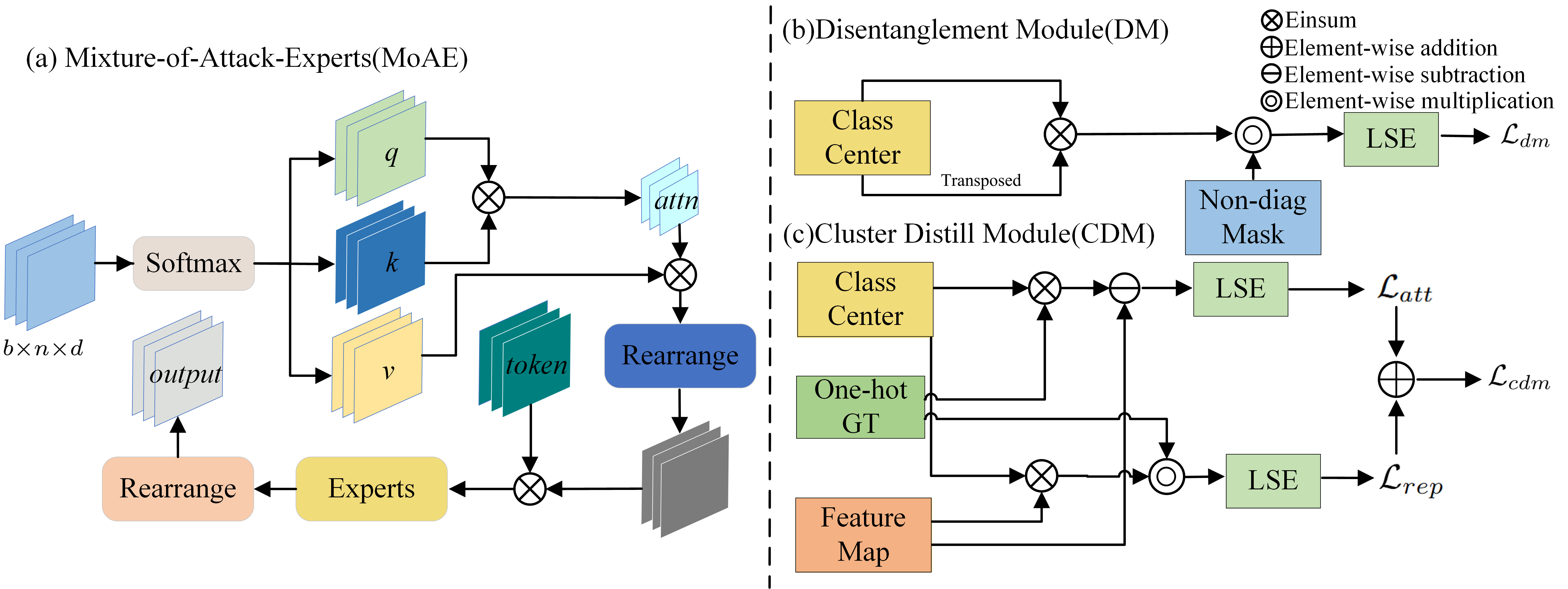} % 替换为图片的路径
    \caption{The detailed implementation structures of the Mixture-of-Attack-Experts (MoAEs), Disentanglement Module (DM), and Cluster Distill Module (CDM) are as follows. The MoAEs build upon the Soft MoEs by incorporating a multi-head attention mechanism to enhance feature processing. The DM utilizes a relationship matrix based on class centers to increase the distance between the centers of different classes. The CDM leverages this relationship matrix to bring each feature closer to its corresponding class center while distancing it from other class centers. Both DM and CDM employ the Log-Sum-Exp (LSE) function to prioritize more distant features.}
    \label{fig3333}
\end{figure*}

As depicted in Fig.~\ref{fig3333}, the DM module is computed using a relationship matrix based on class centers.Assume the input features $X\in\mathbb{R}^{n\times p}$ and labels $Y\in\mathbb{R} ^{n}$. Then we generate one-hot labels $Y_{\mathrm{onehot}}\in\mathbb{R}^{n\times2}$. By utilizing simple summation and statistical methods, the masks $\text{class mask}\in\mathbb{R}^{2\times n}$ for the two classes of live faces and fake faces can be calculated. The formula for the class center is as follows:
\begin{equation}R_\text{center}=\frac{R_\text{s}}{N},\end{equation}
where $N$ is the number of non-zero elements for each class, $R_\text{s}$ is the sum of features for each class. $N$ retains this meaning throughout the text.

The DM is utilized to compute the relative differences in features between classes, thereby promoting the distinct separation of features across different classes. To achieve this, we obtain the inter-class relation matrix $R_\text{c}$ for both live and fake faces:
\begin{equation}R_\text{c}=\frac{{R_\text{center}}\cdot R_\text{center}^T}{\sqrt{p}},R_\text{c}\in\mathbb{R}^{2\times2}.
\end{equation}

Then, we remove diagonal elements $\mathrm{diag}=I_{2\times2}$ and compute the relation $R_\text{or}$ between other classes:
\begin{equation}R_\text{or}=R_\text{c}\cdot(1-\mathrm{diag}).\end{equation}

Building on this basis, we apply the threshold to calculate the relation difference $Q$:
\begin{equation}Q=\max(R_\text{or}-\text{t},0),\end{equation}
where the t is a hyperparameter, which we set to 0.5.

Thus, the DM Center-to-center Loss  $\mathcal{L}_{dm}$ is defined by:
\begin{equation}
\mathcal{L}_{dm} = - \frac{1}{N} \sum_{i=1}^{N} \left( \log \sum_{i \ne j} \exp \left( Q_{i,j} - \max Q_i \right) \right)^2,
\end{equation}
where $max Q_i$ is the maximum value of  $Q_{i,j}$ across its last feature dimensions.It can be seen that we use Log-Sum-Exp here. Due to the properties of Log-Sum-Exp, the actual loss will pull and push all features in the batch, but with varying intensities depending on their relative hardness. We fully considered the impact of distance and subtracted the maximum value to reduce data fluctuations caused by Log-Sum-Exp, ensuring numerical stability.
\subsubsection{Cluster Distillation Module.}Although the DM module separates the centers of the two classes, there might still be instances where live and fake faces are not completely classified based on specific features. To address this, we further designed the CDM (Cluster Discrepancy Minimization). The goal of CDM is to achieve more compact intra-class clustering and greater inter-class separation of features. Specifically, we identify the centers of the live and fake face classes and ensure that each feature in a batch is close to its respective class center while being distant from the center of the other class. This is achieved by designing an attraction loss $\mathcal{L}_{att}$ and a repulsion loss  $\mathcal{L}_{rep}$.
Based on the class centers calculated in Eq. 4, the differences between the features and the mean features are computed. These differences reflect the distance between the features and the class centers:
\begin{equation}{R'}=|\mathrm{X}-R_\text{center} '|,R'\in\mathbb{R}^{n\times p},\end{equation}
where $R_\text{center} '$ is the corresponding class center.
\begin{equation}
\mathcal{L}_{att} = - \frac{1}{N} \sum_{i=1}^{N} \left( \log \sum_{i} \exp \left( R' _{i,j}- \max{R'_i} \right) \right)^2,
\end{equation}
where similar to Eq. 8, $\max{R'_i}$ is the maximum value of  $R'_{i,j}$ across its last feature dimensions.We similarly employ Log-Sum-Exp to emphasize the importance of distance and subtract the maximum value to stabilize the numerical calculations.

The relationship between the features and the class centers can be determined as follows:
\begin{equation} {R'' }=\text{softmax}(X\cdot {R}_\text{center}^T) 
,R'' \in\mathbb{R}^{n\times 2},\end{equation}
\begin{equation}
\mathcal{L}_{rep} = - \frac{1}{N} \sum_{i=1}^{N} \left( \log \sum_{j \ne i} \exp \left( {R''_{i,j}  } - \max {R'' _i } \right) \right)^2.
\end{equation}
We define $\mathcal{L}_{cdm}$ as:
\begin{equation}\mathcal{L}_{cdm}=\mathcal{L}_{att}+\mathcal{L}_{rep}.\end{equation}
\subsubsection{Model Training and Inference.}We calculate the cross-entropy loss using the visual and textual features processed by the CLIP model. In the training phase, we update the parameters of the image encoder with MoAEs and text encoder. The full training objective of MoAE-CR is:
\begin{equation}\mathcal{L}_{total}=\mathcal{L}_{ce}+\mathcal{L}_{dm}+\mathcal{L}_{cdm}.\end{equation}
In the inference stage, the image encoder with MoAEs will adaptively engage different experts based on each example instance. Simultaneously, our DM and CDM, through constraints applied during training, ensure that the image processor achieves intra-class aggregation and inter-class separation, thereby better distinguishing between live and fake faces.
\section{Experiments}
To evaluate the performance of our proposed method in comparison to existing methods, we employed two publicly available datasets, UniAttackData~\cite{fang2024unified} and JFSFDB~\cite{yu2024benchmarking}, for face forgery detection. UniAttackData serves as the primary evaluation dataset due to its advantage of ID consistency. Our method exhibited superior performance and generalization capabilities across both datasets. Additionally, to substantiate the effectiveness of each proposed module, we conducted comprehensive ablation studies.
\subsection{Experimental Settings}

\subsubsection{Datasets.}
UniAttackData extends the CASIA-SURF CeFA dataset by including digital forgery techniques, featuring 1,800 subjects from three ethnic groups and two physical attacks (Print and Replay), with each subject facing 12 digital attacks from six editing and six adversarial methods. It contains 28,706 videos, offering a broader attack variety per identity compared to GrandFake and JFSFDB. Two protocols are defined: Protocol 1 evaluates unified attack detection with all attack types in training, validation, and test sets, while Protocol 2 tests generalization to unseen attacks using a 'leave-one-type-out' approach, divided into P2.1 (unseen physical attacks) and P2.2 (unseen digital attacks). The dataset provides a robust framework for developing and evaluating advanced attack detection methods.

In addition to the UniAttackData, the JFSFDB~\cite{yu2024benchmarking} dataset, introduced by Yu et al., integrates nine subsets. %These include SiW~\cite{liu2018learning}, 3DMAD~\cite{erdogmus2014spoofing}, HKBU-MarsV2~\cite{liu20163d}, MSU-MFSD~\cite{wen2015face}, 3DMask~\cite{yu2020fas}, and ROSE-Youtu for Presentation Attack Detection (PAD), as well as FaceForensics++~\cite{rossler2019faceforensics++}, DFDC~\cite{dolhansky2019deepfake}, and CelebDFv2~\cite{li2020celeb} for Deepfake Attack Detection (DAD). 
The dataset provides two main protocols: separate training, where models address Presentation Attack (PA) and Deepfake Attack (DA) tasks independently, and joint training, which allows simultaneous handling of both tasks. In our study, we employ both protocols to evaluate the effectiveness of our method.

\subsubsection{Implementation Details.}
We configured ViT-B/16 as the image encoder, with the number of experts and heads in the MoAE set to 4 and 2, respectively. The Adam optimizer was employed, with a learning rate of 1e-6 and a weight decay of 5e-4. The model was trained for 300 iterations.
\begin{table}[t]
\centering
\resizebox{\columnwidth}{!}{
\begin{tabular}{c|c|cccc}
\hline
    Prot.     & Method                 & ACER(\%)↓          & ACC(\%)↑            & AUC(\%)↑            & EER(\%)↓           \\ \hline
          & ResNet50               & 1.35                & 98.83               & 99.79               & 1.18               \\
          & VIT-B/16               & 5.92                & 92.29               & 97.00               & 9.14               \\
          & Auxiliary              & 1.13                & 98.68               & 99.82               & 1.23               \\
1           & CDCN                   & 1.40                & 98.57               & 99.52               & 1.42               \\
        & FFD                    & 2.01                & 97.97               & 99.57               & 2.01               \\
          & UniAttackDetection     & 0.52                & 99.45               & 99.96               & 0.53               \\
\textbf{} & \textbf{MoAE-CR(Our)} & \textbf{0.37}       & \textbf{99.47}      & \textbf{99.97}      & \textbf{0.49}      \\ \hline
          & ResNet50               & 34.60±5.31          & 53.69±6.39          & 87.89±6.11          & 19.48±9.10         \\
          & VIT-B/16               & 33.69±9.33          & 52.43±25.88         & 83.77±2.35          & 25.94±0.88         \\
          & Auxiliary              & 42.98±6.77          & 37.71±26.45         & 76.27±12.06         & 32.66±7.91         \\
2          & CDCN                   & 34.33±0.66          & 53.10±12.70         & 77.46±17.56         & 29.17±14.47        \\
         & FFD                    & 44.20±1.32          & 40.43±14.88         & 80.97±2.86          & 26.18±2.77         \\
          & UniAttackDetection     & 22.42±10.57         & 67.35±23.22         & 91.97±4.55          & 15.72±3.08         \\
          
          & \textbf{MoAE-CR(Our)} & \textbf{15.13±12.10} & \textbf{85.41±6.85} & \textbf{92.09±7.11} & \textbf{13.81±8.71} \\ \hline
\end{tabular}
}
\caption{The results of intra-testing on two protocols of UniAttackData, where the performance of Protocol 2 quantified as the mean±std measure derived from Protocol 2.1 and Protocol 2.2.}
\label{table111}
\end{table}

\begin{table}[t]
\centering
\resizebox{\columnwidth}{!}{
\begin{tabular}{c|c|cccc}
\hline
    Prot.     & Method                 & ACER(\%)↓          & ACC(\%)↑            & AUC(\%)↑            & EER(\%)↓           \\ \hline
    & SupContrastive         & 0.64                & 99.32               & 99.95          & 0.68                     \\
    & N-pair         & 1.78               & 98.62              &99.75          & 1.38                     \\
1          & Triplet                & 0.67                & 98.95               & 99.76               & 1.04                \\
          & Hard Triplet                & 1.36                & 98.90               & 99.90               & 1.09                \\
\textbf{} & \textbf{MoAE-CR(Our)} & \textbf{0.37}       & \textbf{99.47}      & \textbf{99.97}      & \textbf{0.49}      \\ \hline
        & SupContrastive         & 16.44±14.14         &68.63±15.08          & 84.77±14.91         & 24.31±22.13               \\
        & N-pair         & 18.79±13.68               & 79.80±12.34               & 85.05±14.14          & 18.66±13.88                   \\
2          & Triplet                & 20.00±14.80         & 69.44±15.93         & 82.52±15.64    &  25.86±20.30            \\
          & Hard Triplet                & 19.56±15.43         & 75.07±11.64         & 88.39±4.13    &  16.91±7.09            \\
          & \textbf{MoAE-CR(Our)} & \textbf{15.13±12.10} & \textbf{85.41±6.85} & \textbf{92.09±7.11} & \textbf{13.81±8.71} \\ \hline
\end{tabular}
}
\caption{The results of intra-testing on the two protocols of UniAttackData with different losses, where the performance of Protocol 2 quantified as the mean±std measure derived from Protocol 2.1 and Protocol 2.2.}
\label{table222}
\end{table}
\begin{table}[t]
\centering
\resizebox{\columnwidth}{!}{
\begin{tabular}{c|cccc}
\hline
Method        & ACER(\%)↓ & ACC(\%)↑ & AUC(\%)↑ & EER(\%)↓ \\
\hline
ResNet50      & 7.70        & 90.43    & 98.04    & 6.71     \\
VIT-B/16      & 8.75       & 90.11    & 98.16    & 7.54     \\
Auxiliary     & 11.16      & 87.40     & 97.39    & 9.16     \\
CDCN          & 12.31      & 86.18    & 95.93    & 10.29    \\
FFD           & 9.86       & 89.41    & 95.48    & 9.98     \\
\textbf{MoAE-CR(Our)} & \textbf{4.40}       & \textbf{95.33}    & \textbf{98.97}    & \textbf{4.66}     \\
\hline
\end{tabular}}
\caption{This table presents the results on the JFSFDB dataset under the p2 intra protocol, where our proposed MoAE-CR has the SoTA performance. }
\label{table333}
\end{table}

\subsection{Performance}
To evaluate the proposed algorithm in Unified Attack Detection (UAD), we use standard metrics from physical and digital forgery detection: average classification error rate (ACER), overall detection accuracy (ACC), area under the curve (AUC), and equal error rate (EER). ACER and ACC are calculated based on thresholds from the development set.

Additionally, the robustness of our method is demonstrated through comparisons with a range of established competitors in face anti-spoofing and backbone networks, including ResNet50, ViT-B/16, FFD~\cite{dang2020detection}, CDCN~\cite{yu2020fas}, Auxiliary (Depth)~\cite{liu2018learning}, and UniAttackDetection~\cite{fang2024unified}. 

Table~\ref{table111} shows that our method, MoAE-CR, achieves state-of-the-art performance across all metrics (ACER, ACC, AUC, EER) on UniAttackData. Its strong results in Protocol 2 demonstrate excellent generalization to "unseen" attacks. 

In Protocol 1, our method achieves an ACER of 0.37\%, surpassing the previous best of 0.52\%. The ACC reaches 99.47\%, exceeding the prior best of 99.45\%. Both AUC and EER also show notable improvements.

In Protocol 2, our method demonstrates even more significant advances, with an average ACER of 15.13\% and an average ACC of 85.41\%, far outperforming previous methods. AUC and EER similarly show substantial improvements.

To validate the effectiveness of our DM and CDM modules, we replaced them with conventional techniques such as triplet loss and supervised contrastive loss. As shown in Table~\ref{table222}, our DM and CDM modules outperform these mainstream methods in both protocols, highlighting their superiority in handling combined digital and physical attack tasks.

To further assess the efficacy of the MoAE-CR method, we conducted additional tests on the JFSFDB dataset. As shown in Table~\ref{table333}, our proposed method, MoAE-CR, achieves SoTA performance with an ACER of 4.40\%, an ACC of 95.33\%, an AUC of 98.97\%, and an EER of 4.66\%. These results further substantiate the superior performance of our method compared to previous works.

\subsection{Ablation Study}
\begin{table}[t]
\centering
\resizebox{\columnwidth}{!}{
\begin{tabular}{cccccccc}
\toprule
\textbf{CLIP} & \textbf{SoftMoE} & \textbf{MoAE} & \textbf{DM} & \textbf{CDM} & \textbf{ACER(\%)}$\downarrow$ & \textbf{ACC(\%)}$\uparrow$ &\textbf{AUC(\%)}$\uparrow$  \\ 
\midrule
\checkmark  & - & - & - & - &0.79 &98.91  &99.76  \\
\checkmark & \checkmark & - & - & - & 0.66 &99.01  &99.82\\
\checkmark & - & \checkmark & - & - &0.49  &99.24  &99.79\\
\checkmark & - & \checkmark & \checkmark &- &0.95  &98.73 &99.79\\
\checkmark & - & \checkmark & - & \checkmark &0.54 &99.28 &99.86\\
\checkmark & - & \checkmark & \checkmark & \checkmark &\textbf{0.37}  &\textbf{99.47}  &\textbf{99.97}\\
\bottomrule
\end{tabular}}
\caption{Ablation of each component was conducted on the UniAttackData under Protocol I.}
\label{table444}
\end{table}
\subsubsection{Effectiveness of Each Component.}
To assess the contribution of each component in our framework, we conducted ablation studies beginning with the baseline framework, CLIP. Specifically, Soft MoEs were employed to process different features through distinct experts and merge them via a soft routing mechanism, as shown in Table~\ref{table444}, leading to performance changes of -0.13\% (ACER), +0.1\% (ACC), and +0.06\% (AUC).

Upon introducing fine-grained improvements to Soft MoEs, the optimized results for the three metrics improved to 0.49\% (ACER), 99.24\% (ACC), and 99.79\% (AUC). This suggests that incorporating a multi-head attention mechanism within Soft MoEs further enhances the model's ability to represent features. While the individual introduction of DM and CDM into the Fine-Grained MoEs framework resulted in slight decreases across each metric, their combined application produced optimal outcomes of 0.37\% (ACER), 99.47\% (ACC), and 99.97\% (AUC). These results indicate that DM and CDM exhibit a positive synergistic effect, and their integration delivers significant improvements.

\subsubsection{Effects of the Number of Experts and Heads.}
\begin{table}[t]
\centering
\resizebox{0.6\columnwidth}{!}{ % Adjust the width as needed
\begin{tabular}{c|c c c}
\hline
ACER(\%)↓        & \multicolumn{3}{c}{Num of heads} \\
\hline
Experts & $ \times 2 $ & $ \times 4 $ & $ \times 8 $ \\
\hline
$ \times 2 $       & 0.90        & 0.41       & 0.71 \\
$ \times 4 $       & \textbf{0.37}       & 0.41        & 0.52 \\
$ \times 8 $       & 0.67       & 1.09       & 1.60 \\
\hline
\end{tabular}}
\caption{This table shows the effects of the number of experts and attention heads.}
\label{table555}
\end{table}
We evaluated the impact of the number of heads and experts in MoAE. As shown in Table 5, the model generally performs better with four experts. However, increasing the number of attention heads leads to a decline in performance, suggesting that excessive focus can negatively impact the model's effectiveness. Similarly, having too many experts results in performance degradation, indicating that an excessive number of experts may cause overfitting or optimization difficulties. Therefore, we recommend using two attention heads and four experts as a better trade-off between performance and efficiency.

\subsection{Visualization and Analysis}
\begin{figure}[t]
    \centering
    \includegraphics[width=\columnwidth]{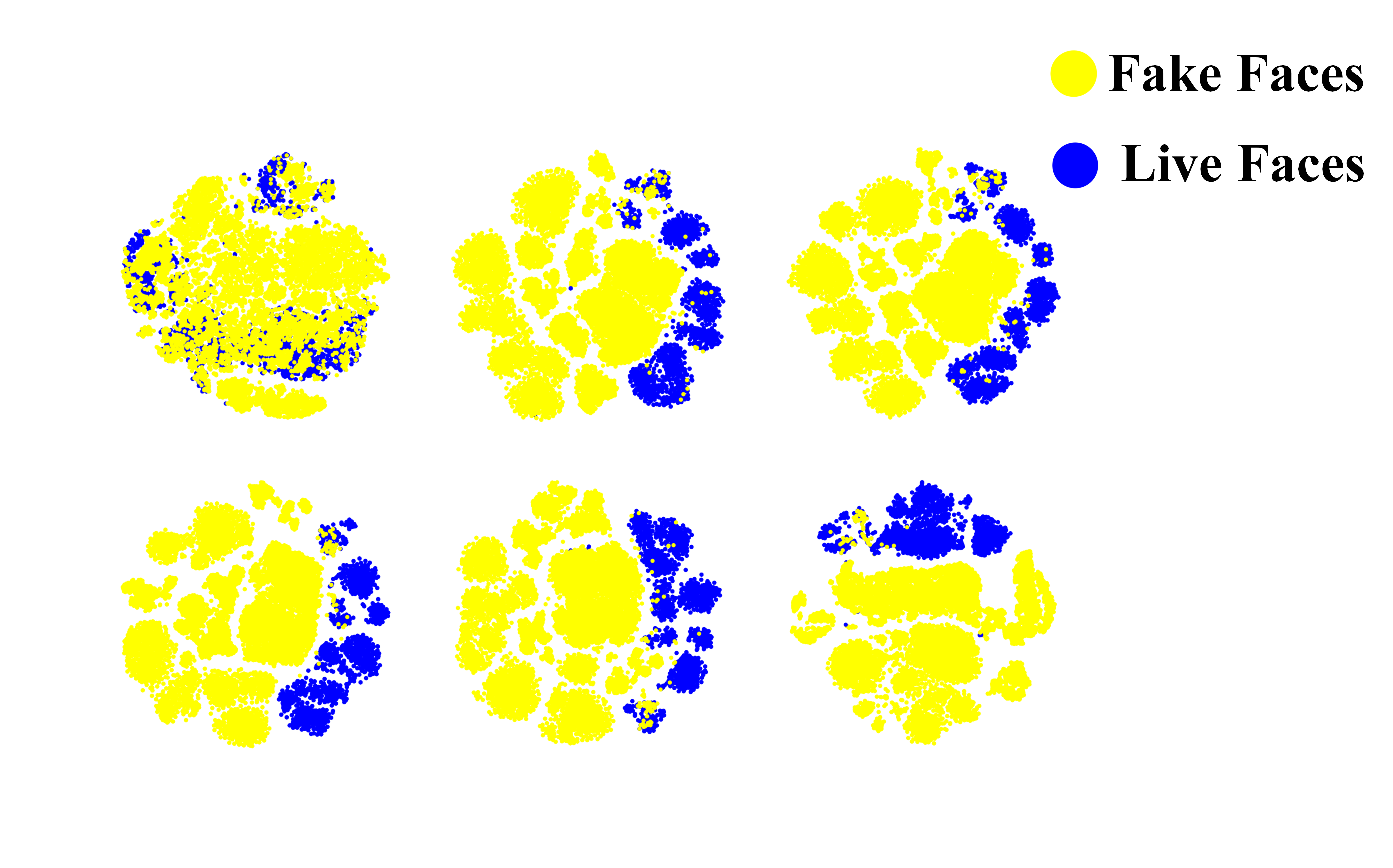} % 替换为图片的路径
    \caption{The figure presents the feature distribution visualization analysis of UniAttackData using the following methods: vanilla CLIP (top left), CLIP with SoftMoE (top center), CLIP with MoAE (top right), MoAE with DM (bottom left), MoAE with CDM (bottom center), and MoAE-CR (bottom right).}
    \label{fig555}
\end{figure}
Using t-SNE and Matplotlib for feature visualization, as shown in~\ref{fig555}, the feature distribution significantly improves upon introducing SoftMoE. The proposed MoAE-CR further optimizes the feature distribution compared to SoftMoE. The introduction of DM or CDM constraints enhances within-class clustering and improves class separation. The most notable improvements are observed when both DM and CDM constraints are applied simultaneously. This combined approach leads to clearer inter-class separation and more compact within-class distributions, facilitating the identification of the decision boundary. Further experimental analyses can be found in the supplementary materials.

\section{Conclusion}
In this work, we introduce the MoAE-CR framework to effectively tackle the challenges arising from combined digital and physical attacks, addressing these at both the feature and loss levels. At the feature level, our framework integrates Soft MoEs, and we further propose MoAE to enhance feature processing. At the loss level, we incorporate two constraint modules, DM and CDM, to facilitate more accurate classification by promoting a more balanced distribution between live and fake faces. Extensive experiments and visual analyses substantiate the superiority of the proposed MoAE-CR framework.

\section{Acknowledgments}
This work was supported by Beijing Natural Science Foundation JQ23016, the Chinese National Natural Science Foundation Projects 62476273, 62406320, U23B2054, 62276254, the Science and Technology Development Fund of Macau Project 0044/2024/AGJ, 0123/2022/A3, 0070/2020/AMJ, 0096/2023/RIA2, Ant Group and InnoHK program.
\bibliography{aaai25}
\end{document}